%% file: main.tex
\documentclass{article} 
\usepackage[preprint]{colm2025_conference}

\usepackage{microtype}
\usepackage{hyperref}
\usepackage{url}
\usepackage{booktabs}

\usepackage{lineno}

\definecolor{darkblue}{rgb}{0, 0, 0.5}
\hypersetup{colorlinks=true, citecolor=darkblue, linkcolor=darkblue, urlcolor=darkblue}

\usepackage{graphicx}
\usepackage{caption}
\usepackage{subcaption}
\usepackage{amsmath}
\usepackage{booktabs}
\usepackage{color,soul}
\usepackage{xcolor}
\usepackage{pifont}
\usepackage{amssymb}
\usepackage{enumitem}
\usepackage{mathtools}
\usepackage{multirow}
\usepackage{float}
\usepackage{xspace}
\usepackage{tcolorbox}   
\usepackage{wrapfig}
\usepackage{tabularx}
\usepackage{colortbl}
\usepackage[capitalize,noabbrev]{cleveref}

\def\ie{{\em i.e., }\xspace}

\newcommand{\hlc}[2]{\colorbox{#1}{\strut #2}}

\title{Noiser: Bounded Input Perturbations for \\ Attributing Large Language Models}

\author{
    \textbf{Mohammad Reza Ghasemi Madani}\textsuperscript{1}\thanks{Correspondence to: \texttt{mr.ghasemimadani@unitn.it}.} \;
    \textbf{Aryo Pradipta Gema}\textsuperscript{2} \;
    \textbf{Gabriele Sarti}\textsuperscript{3}\\
    \textbf{Yu Zhao}\textsuperscript{2} \;
    \textbf{Pasquale Minervini}\textsuperscript{2,4} \;
    \textbf{Andrea Passerini}\textsuperscript{1} \\
    \textsuperscript{1}University of Trento \quad
    \textsuperscript{2}University of Edinburgh \quad
    \textsuperscript{3}CLCG, University of Groningen \\
    \textsuperscript{4} Miniml.AI
}

\begin{document}

\ifcolmsubmission
\linenumbers
\fi

\maketitle



\begin{abstract}

Feature attribution (FA) methods are common post-hoc approaches that explain how Large Language Models (LLMs) make predictions.
Accordingly, generating faithful attributions that reflect the actual inner behavior of the model is crucial.
%
In this paper, we introduce \textsc{\textbf{Noiser}}, a perturbation-based FA method that imposes bounded noise on each input embedding and measures the robustness of the model against partially noised input to obtain the input attributions.
Additionally, we propose an \emph{answerability} metric that employs an instructed judge model to assess the extent to which highly scored tokens suffice to recover the predicted output.
%
%
%
Through a comprehensive evaluation across six LLMs and three tasks, we demonstrate that Noiser consistently outperforms existing gradient-based, attention-based, and perturbation-based FA methods in terms of both faithfulness and answerability, making it a robust and effective approach for explaining language model predictions.
\end{abstract}

\section{Introduction}

Transformer-based language models \citep{vaswani2023attention} are fundamental to the latest advancements in natural language processing \citep{gemmateam2024gemmaopenmodelsbased, touvron2023llama, qwen, deepseekai2025deepseekr1, openai2024gpt4technicalreport}. 
However, they are often perceived as opaque \citep{rudin2019stop,doshivelez2017rigorous,lipton2018mythos}, sparking significant interest in the development of algorithms that can automatically explain the behavior of these models \citep{denil2015extraction, sundararajan2017axiomatic, Camburu_NEURIPS2018, rajani-etal-2019-explain, luo2022local}.

Feature attribution (FA) techniques are popular post-hoc methods that generate token-level importance scores to highlight the contribution of each token to a prediction \citep{denil2015inputxgradient, jain2020attention, kersten-etal-2021-attention}. 
The top-$k$ important tokens are typically considered as the prediction rationale \citep{zaidan-etal-2007-using, sundararajan2017integratedgradients, deyoung-etal-2020-eraser}. 
%
The quality of a rationale is often evaluated using \emph{faithfulness} metrics, which measure to what extent the rationales accurately reflect the downstream task on model predictions.
%

Perturbation-based FA methods aim to explore neural networks by modifying the input of a model and observing the changes in the output to indicate which parts of the input are particularly important for inference. 
These methods are widely adopted in computer vision, leveraging the continuous nature of image inputs, where localized noise or masking preserves semantic coherence and avoids distribution shifts \citep{IVANOVS2021228}.
In contrast, NLP models face inherent challenges due to the discrete structure of the text, where even minor perturbations—whether token substitutions or embedding modifications—can push inputs out-of-distribution (OOD), destabilizing predictions and confounding attribution analysis \citep{10.1109/TVCG.2018.2865230}. 

This divergence underscores the need for bounded perturbations in NLP, ensuring perturbed inputs remain in-distribution. Our work bridges this gap by exploring noise thresholds that 
alter token embeddings while preserving the original prediction to limit perturbation-induced OOD issues. Particularly, we introduce a perturbation-based FA by exploring a model's robustness against \emph{noisy inputs}—examples created by introducing small alterations to the input embeddings without changing the model's original prediction—enabling reliable explanations grounded in the model’s trained operational domain while quantifying feature importance through robustness to controlled perturbations. Our work makes the following contributions:


\begin{itemize}[noitemsep, leftmargin=*]
    \item We empirically show that \textsc{Noiser} is consistently more faithful than nine popular FAs by conducting comprehensive experiments, covering three tasks and six LMs of varying sizes from three different model families;
    \item We propose a new plausibility metric, \emph{answerability}, which measures the extent to which the top-$k\%$ attributed input tokens sufficiently support the target output. By leveraging language models, this metric assesses whether a minimal subset of input tokens is adequate for generating the expected prediction, providing a simulatable alternative to human plausibility judgments.
\end{itemize}

\section{Background}

\subsection{Generative Language Modeling}
In generative language modeling, the input consists of a sequence of tokens, denoted as $\text{X}=[x_0, \ldots, x_{T-1}]$. The objective is to develop a model, $\mathcal{F}\theta$, that estimates the probability distribution $P$ over the token sequence $\text{X}$. In this context, $\mathcal{F}\theta$ represents a specific pre-trained generative language model characterized by parameters $\theta$.
\begin{equation*}
    P(x_0, \ldots, x_{T-1}) = \mathcal{F}_\theta(x_0) \prod_{t=1}^{T-1} \mathcal{F}_\theta(x_t \mid x_0, \ldots, x_{t-1})
\end{equation*}

\subsection{Input Importance for Generative LMs}
Given a model $\mathcal{F}_\theta$, our objective is to determine the importance distribution of the input tokens for each predicted token $x_T$, based on the preceding sequence $\text{X} = [x_0, \ldots, x_{T-1}]$. A feature attribution method, denoted as $e_T$, applied at position $T$, yields an importance distribution $\text{S}_T = [s_0, \ldots, s_{T-1}]$ corresponding to the target token $x_T$, where a higher value of $s_i$ indicates greater importance of the input token $x_i$ in predicting $x_T$. 
\begin{equation*}
    e_T(\mathcal{F}_\theta, \text{X}, x_T) \to \text{S}_T
\end{equation*}

\subsection{Bounded Perturbations}
Bounded perturbations refer to small, structured uncertainties in mathematical systems where a specified constraint limits the perturbation magnitude. These are critical for analyzing system robustness against disturbances while ensuring predictable behavior.

Perturbation is a minor alteration to a system, such as $\delta$ added to a nominal matrix $A$, resulting in $A + \delta$. This captures uncertainties or disturbances. If the perturbation magnitude is bounded as $\|\delta\| \leq \epsilon$, where $\epsilon > 0$, it is called a bounded perturbation. 
Consider the nominal system $\dot{x} = Ax$. Under a perturbation $\delta$, the system becomes:
\begin{equation*}
    \dot{x} = (A + \delta)x.
\end{equation*}



\section{Our Method}
\label{sec:our-method}
%
Let $\mathbf{n} \in \mathbb{R}^{d_{\text{model}}}$ denote a noise vector where each component $n_i \sim \mathcal{N}(0, 1)$. 
We form noisy examples from original inputs by imposing small perturbations to the input embeddings, such that the noisy input results in the model outputting an incorrect answer.
For this purpose, we first pass a prompt $\text{X}=[x_0, \ldots, x_{T-1}]$ into $\mathcal{F}_\theta$ to collect the probability distribution, $P$, over the model's vocabulary with $x_T$ being the most likely output (\ie{$\mathcal{F}_\theta(\text{X}) = x_T$}).

In the next step, we utilize a \textit{binary search} algorithm to find the \textbf{maximum} scaling factor $k$ such that if we perturb the embedding of a targeted token with $\mathbf{n}_{\text{scaled}} = k \cdot \mathbf{n}$ the model wouldn't change its initial prediction $x_T$. 
Specifically, we set $x_i:= x_i + \mathbf{n}_{\text{scaled}}$ and let $\mathcal{F}_\theta$ to continue, giving us a set of corrupted probabilities $P^*_{x_i}$. 
Because $\mathcal{F}_\theta$ partially loses information about the corrupted token, the probability of $x_T$ from the first step would likely be lower in $P^*_{x_i}$. 

We repeat the process for each token until we obtain $\text{K} = [k_0, \ldots, k_{T-1}]$ where each $k_i$ is the maximum scaling factor such that if we corrupt the embeddings of $x_i$ using  $\mathbf{n}_{\text{scaled}} = k_i \cdot \mathbf{n}$, the model wouldn't change its original output. The mathematical representation of $K$ is illustrated below:

\begin{equation*}
\label{eq:scale}
   K = \{k_i \mid \forall k>k_i \Rightarrow 
   \mathcal{F}_\theta(\text{X}_{\text{perturbed} \mid k}) \neq x_T, \mathcal{F}_\theta(\text{X}_{\text{perturbed} \mid k_i})=x_T\}, \quad i \in \{0, \ldots, t-1\}
\end{equation*}

where $\text{X}_{\text{perturbed}\mid k} = [x_0 \ldots (x_i + \mathbf{n}_\text{scaled}) \ldots x_{t-1}]$ is the input sequence in which $x_i$ is altered with $\mathbf{n}_\text{scaled} = k \cdot \mathbf{n}_\text{bounded}$.
The equation above indicates that each scale factor $k_i$ is such that for all values $k$ greater than $k_i$ if we perturb $x_i$ using $\mathbf{n}_\text{sclaed} = k \cdot \mathbf{n}_\text{bounded}$ to create a noisy input $\text{X}_{\text{perturbed}\mid k}$, $\mathcal{F}_\theta$ would return a different output from the original one ($x_T$).

In the final step, we find the $k_{\min} = \min(K)$ to generate the final noise samples $\mathbf{n}_\text{scaled} = k_\text{min} \cdot \mathbf{n}$ to add to  each token embedding and obtain the token scores using the following:

\begin{equation*}
\label{eq:score}
    S = \{s_i \mid s_i = p(\text{X}) - p(\text{X}_{\text{perturbed} \mid k_{\min}})\}, \quad i \in \{0, \ldots, t-1\}
\end{equation*}


Using $k_\text{min}$, we ensure to perturb the input enough to reach a flipping point in prediction to get the minimal set of features needed to achieve this outcome. 
The intuition is that tokens with higher importance are more sensitive to noise injection, resulting in a larger reduction in the model's output likelihood.


To show the effectiveness of selecting $k_{\min}$, we propose different boundings and measure their faithfulness. We analyse \textbf{i)} using maximum noise across tokens ($k_{\max}$); \textbf{ii)} individual token maximum noise where $k$ is different for each input token and is the maximum the model can tolerate ($k_{\max}$ per token); \textbf{iii)} norm-bounded setting where the noise vector $\mathbf{n}$ is divided by the expected value of the noise vector $L_p$ norm, $\mathbb{E}\left[\|\mathbf{n}\|_p\right]$; and \textbf{iv)} random $k$ where $k$ is randomly selected from the uniform distribution. The details of each configuration are provided in \cref{sec:res}.

\section{Experiment}

\subsection{Model \& Data}
In our study, we employ variants of Qwen~\citep{qwen}, Gemma~\citep{gemmateam2024gemmaopenmodelsbased}, and Llama~\citep{touvron2023llama} models.
We choose our models to span from hundreds of millions to a few billion parameters as we want to explore how the model size affects the faithfulness of each FA. All models used are publicly available\footnote{We use checkpoints from the Huggingface library for each model.}.

We use \textsc{Known} dataset\footnote{Dataset can be found at: \url{https://rome.baulab.info/data/dsets/known_1000.json}} provided by \citet{meng2023locating} and \textsc{Long-Range Agreement} \citep[\textsc{LongRA};][]{vafa2021rationalessequential} to conduct our analysis. Besides, for long generation we utilize \textsc{WikiBio} \citep{lebret2016neuraltext}.
The following is an instance from the \textsc{Known} dataset.

\begin{quote}
    \textit{LeBron James professionally plays the sport of} [\textit{\textbf{basketball}}]
\end{quote}

The LongRA dataset consists of word pairs that exhibit either a semantic or syntactic relationship. Additionally, \citet{vafa2021rationalessequential} incorporate a distractor sentence, which provides no relevant information about the word pair, to evaluate long-range agreement. An example from the LongRA dataset is shown below, with the distractor included in parentheses.

\begin{quote}
    \textit{When my flight landed in \textbf{Japan}, I converted my currency and slowly fell asleep. (I had a terrifying dream about my grandmother, but that’s a story for another time). I was staying in the \textbf{capital},} [\textbf{\textit{Tokyo}}]
\end{quote}

\textsc{WikiBio} is a dataset consisting of Wikipedia biographies. We use the first two sentences as a prompt, similar to \cite{manakul2023selfcheckgpt}. The model is then expected to continue generating the biography. This task is inherently more open-ended compared to the previous two.
%

\begin{quote}
    \textit{Super Mario Land is a 1989 side-scrolling platform video game}
\end{quote}

The computation of each task's faithfulness is provided in \cref{sec:details}.

\subsection{Baselines}
Following previous works, we compare our rationalization method to a variety of gradient- and attention-based baselines \citep{vafa2021rationalessequential}. 
\textbf{Input$\times$Gradient} \citep{denil2015inputxgradient} uses embedding gradients multiplied by the embeddings;
\textbf{Integrated Gradients} \citep{sundararajan2017integratedgradients} integrate overall gradients using a linear interpolation between a baseline input (all zero embeddings) and the original input.
\textbf{Gradient SHAP} \citep{lundberg2017gradientshap} compute the gradient w.r.t. randomly selected points between the inputs and a baseline distribution;
\textbf{DeepLIFT} \citep{shrikumar2019deeplift} compares the activation of each neuron to its `reference activation' and assigns contribution scores according to the difference.
\textbf{Sequential Integrated Gradients} \citep{enguehard2023sequentialint} extends Integrated Gradients by breaking down the input perturbation into sequential steps, computing gradients at each step, and aggregating them to provide more stable and interpretable attributions,
%
%
\textbf{Last Attention} \citep{jain2020attention} uses the last-layer attention weights averaged across heads;
\textbf{Attention Rollout} \citep{abnar2020rollout} recursively computing the token attention in each layer, e.g., computing the attention from all positions in layer $l_i$ to all positions in layer $l_j$, where $j < i$;
\textbf{LIME} \citep{ribeiro2016lime} trains a linear surrogate model using data points randomly sampled locally around the prediction. 
\textbf{Occlusion} \citep{Zeiler2014occlusion} involves systematically occluding different portions of the input and observing the impact on the output confidence. 

%

\subsection{Faithfulness Metrics}
To assess whether a rationale extracted with a given FA is faithful, \ie{actually reflects the true model reasoning} \citep{jacovi-goldberg-2021-aligning}, various faithfulness metrics have been proposed \citep{Arras_2017, serrano-smith-2019-attention, jain2020attention, deyoung-etal-2020-eraser}. 
Sufficiency and comprehensiveness \citep{deyoung-etal-2020-eraser} are two widely used metrics that effectively capture rationale faithfulness \citep{chrysostomou-aletras-2021-enjoy, chan-etal-2022-comparative}. 
Both metrics use a hard erasure criterion for perturbing the input by entirely removing (\ie{comprehensiveness}) or retaining (\ie{sufficiency}) the rationale to observe changes in predictive likelihood.
This hard criterion ignores the importance of each individual token, treating them all equally for computing sufficiency and comprehensiveness.

We evaluate rationales using soft sufficiency (Soft-NS) and comprehensiveness (Soft-NC) proposed by \citet{zhao-aletras-2023-incorporating} to measure the faithfulness of the full importance distribution. 
Using these metrics, instead of entirely removing or retaining tokens from the input, we randomly mask parts of the token vector representations proportionately to their FA importance. 
The summation of Soft-NC and Soft-NS is considered as the final faithfulness score. For the detailed implementation of these metrics, please refer to \cref{sec:metrics}.

\subsection{Answerability Metrics}
\begin{figure}
    \centering
    \includegraphics[width=0.7\linewidth]{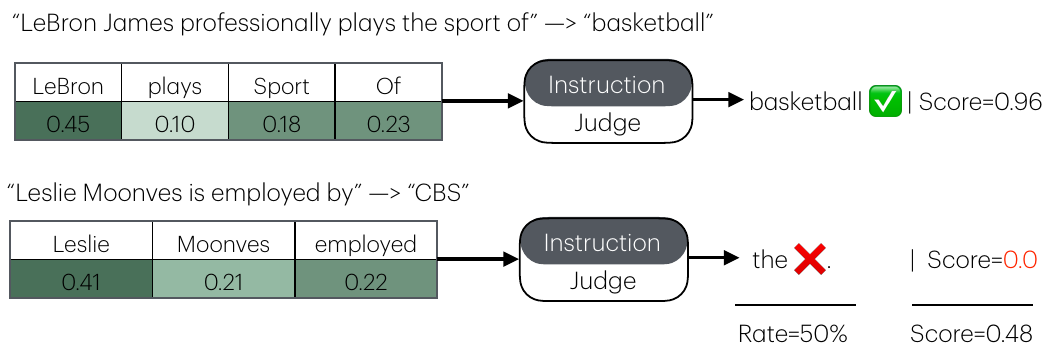}
    \caption{Answerability metrics evaluation. To get the answerability metrics, the judge model is instructed to predict the completion token given a limited set of tokens from the original prompt.}
    \label{fig:answerability}
\end{figure}


The utilization of LLMs has emerged as a prominent trend across numerous research domains \citep{peng2023instructiontuninggpt4, zhou2023limaalignment, alpaca}. With any given instructions, LLMs are expected to generate responses that align with these instructions \citep{chen2024alpagasus, li2024selfalignment, xu2023wizardlm, longpre2023flancollection}.

This capability, known as the ``instruction following'' ability, serves as a key metric for assessing
the effectiveness of LLMs \citep{chen2024alpagasus, zhao2024longalignments, alpaca, zheng2023judgingllm}. To facilitate a more thorough assessment, several benchmarks have been introduced with a focus on instruction following \citep{zhou2023instructionfollowing, qin2024infobench}.


We exploit this progress in our answerability metric by framing attribution evaluation as an instruction-based completion task.
To see whether the attributions illustrate any meaningful association with a predicted output, we extend our evaluation of FAs through prompt engineering. 
For this purpose, we aggregate the attribution scores of each word's sub-tokens to derive word-level scores. 
Then we select the top-$k\%$ most important words w.r.t. their scores and feed these words as input to a judge model along with a task prompt. 
The task prompt asks the judge model to predict the completion token using this limited set of words. 
\citet{feldhus-etal-2023-saliency} offer a complementary perspective—while they leverage LLMs for generating interpretability-enhancing verbalizations, our approach instead uses an LLM to directly quantify whether the selected tokens are sufficient for the prediction task.

We evaluate attributions by computing the number of samples for which the judge model generates the correct output, which we define as the FAs \emph{\textbf{answerability rate}}. 
Additionally, for these correctly predicted samples, we aggregate the word-level attribution scores to obtain the so-called \emph{\textbf{answerability score}}. For both metrics, higher values indicate better performance. In our evaluation, a higher answerability rate means that a larger proportion of samples allow the LM judge to correctly predict the output using only the minimal set of tokens. Likewise, a higher answerability score—reflecting a greater aggregated attribution mass within that token set—indicates that the attribution method is more effective at isolating the minimal semantic requirements for prediction.
%

This evaluation pipeline specifically applies to datasets such as \textsc{Known} where the gold label is a meaningful word that must be inferred from the input sequence.
\cref{fig:answerability} shows an answerability evaluation example.
The prompt used for this evaluation is shown in \cref{plaus_prompt}.

\subsection{Implementation Details} 
\label{sec:details}
None of our experiments involved training or fine-tuning any language models. All FAs are built upon Inseq library \citep{sarti-etal-2023-inseq,sarti-etal-2024-democratizing} except Last Attention and Attention Rollout, which we used the codebase from \cite{zhao2024reagent}.
%
For \textsc{Noiser}, we generate 10 different noise vectors during the corrupted run for more consistent results.
Binary search is done in 10 steps, yielding the accuracy of $\approx0.001$ for the scaling factor ($k$).
For \textsc{Knonw} and \textsc{LongRA} datasets, where the models must provide a single output, we filter down samples to the ones that the model can correctly generate the gold output. See \cref{sec:data-stat} for the details.
For \textsc{WikiBio}, we generate 10 tokens for input completion. To obtain the faithfulness score in this task, we compute the faithfulness of each next token w.r.t. all the previous tokens and consider the averaged score as the final faithfulness.
The judge model used to get the answerability metrics is \texttt{Llama-3.3-70B-Instruct-Turbo}. We chose the top-50\% of the most important words from the input prompt to get the answerability score and rate.

\section{Results}
\label{sec:res}
\input{tables/faithfulness-all}

\Cref{tab:faithfulness-all} presents the faithfulness scores across different tasks. Following \citet{zhao2024reagent}, each score is computed as the logarithm of the ratio between the method’s score and the random baseline. Consequently, scores below zero indicate less faithful methods than the random baseline, \ie{unfaithful}. As shown in \cref{tab:faithfulness-all}, faithfulness varies across different FA methods and generative models. Notably, \textsc{Noiser} consistently achieves higher faithfulness scores across all tasks and models, outperforming traditional FAs. This suggests that \textsc{Noiser} provides more reliable attributions, reinforcing its effectiveness in evaluating model faithfulness.


To demonstrate the effectiveness of selecting $k_{\min}$, we compare its faithfulness performance against alternative bounding strategies across different models in \cref{tab:k-effect}. The results show that $k_{\min}$ consistently yields the highest faithfulness scores, confirming its superiority in preserving model behavior under noise perturbation.

Since $k_{\min}$ is model-dependent, we introduce norm-bounding as a flexible alternative, where the noise vector $\mathbf{n}$ is scaled based on the model’s embedding size (see \Cref{sec:scaling-effect}). 
We further compare our approach with $k_{\max}$, which applies the maximum $k$ across all input tokens ($\max(K)$), and a variant, $k_{\max}$ per token, which applies a per-token maximum scaling factor. The latter performs slightly better, as it results in a less aggressive perturbation than the global $k_{\max}$, reducing the likelihood of extreme changes in model behavior.

Additionally, we analyze the effects of unbounded scaling ($k = 1$) and random $k$, where $k$ is sampled from a uniform distribution for each input sample. The consistently lower faithfulness scores in these settings highlight the necessity of proper bounding strategies to maintain faithfulness.

Overall, $k_{\min}$ is the only configuration that guarantees the model does not change its prediction under noise, making it the most reliable choice. The detailed computation of expected norm values is provided in \Cref{sec:scaling-effect}.


%

\input{tables/k-effect}


The \emph{answerability} rate and score are reported in \cref{tab:answerability}. While \textsc{Noiser} achieves the highest answerability score in most cases and on average, Occlusion attains the highest answerability rate. This indicates that when \textsc{Noiser} attributions are deemed answerable (rate), the importance scores assigned to the top-$k\%$ tokens are significantly high, which is desirable. In contrast, Occlusion produces a higher number of answerable attributions but with lower scores, implying that it does not assign as much weight to key tokens.

To provide a more flexible analysis of answerability metrics, we examine cases where the gold prediction appears within the top-5 predictions of the judge model. Under this evaluation, the gap between \textsc{Noiser}'s answerability score and those of other baselines widens, while its answerability rate also improves and approaches that of Occlusion, which achieves the best rate.

\input{tables/answerability-top1}

\begin{figure}[h]
    \centering
    \includegraphics[width=1\linewidth]{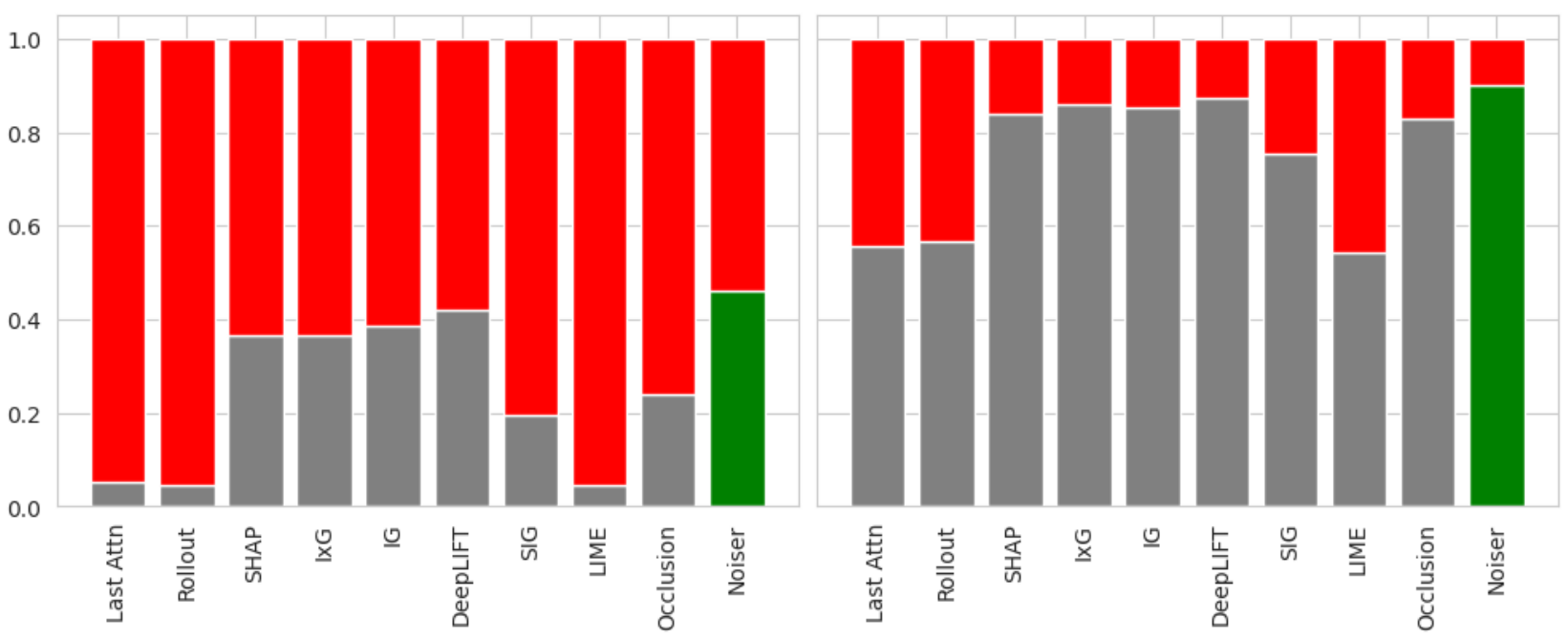}
    \caption{The aggregated score that each FAs put different parts of inputs from the ``capital-world'' subclass in \textsc{LongRA} dataset. The red indicates the score assigned to the undesired part (e.g., distractor). The left image illustrates the aggregated score on ``country''+``capital'' token. The right image indicates the overall score on the main part against the distractor.}
    \label{fig:longra-combined}
\end{figure}

Another aspect regarding the FA methods' efficiency that we evaluate is their ability to identify the minimal set of tokens most relevant to the output. We visualize the importance scores assigned by each method to critical tokens in the \textsc{LongRA} ``country-capital'' category. Additionally, we examine the distribution of importance scores across the distractor and main parts in \cref{fig:longra-combined}. As shown in \cref{fig:longra-combined}, \textsc{Noiser} assigns the highest importance to critical tokens while effectively disregarding the distractor section, demonstrating a stronger focus on the main part compared to the best-performing baselines.


%


Finally, we analyzed the minimum proportion of top attributions required for each FA method to ensure that the judge model correctly predicts the original output. To determine this value, we first computed attributions for each sample using a given FA method. Then, starting with the full set of tokens, we iteratively removed the least important tokens one by one until the judge model produced an incorrect prediction. We repeated this process across all samples, averaging the proportion of retained tokens to obtain the final minimum top-$k\%$ required for accurate prediction. A lower value indicates a more effective FA method in identifying the most relevant attributions. In this regard, Occlusion requires the least number of tokens overall, which aligns with the results in \cref{tab:answerability}, while Lime and \textsc{Noiser} take the second and third best place with minimal difference.

\begin{figure}
    \centering
    \includegraphics[width=0.6\linewidth]{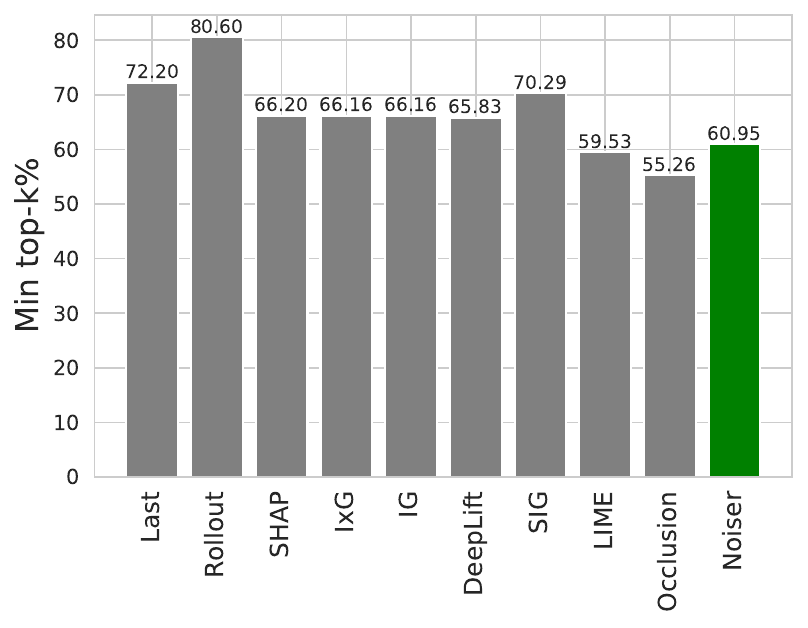}
    \caption{Minimum top-$k\%$ attribution required for the judge model to retain the correct prediction across different feature attribution methods. Lower values indicate higher attribution accuracy, as fewer tokens are needed to maintain the original output.}
    \label{fig:min-k}
\end{figure}

\input{tables/highlights}

\section{Related Works}
Post-hoc explanation methods, such as FA techniques, are applied retrospectively by seeking to extract explanations after the model makes a prediction. 
Most FAs have been proposed in the context of classification tasks, where a sequence input $\text{X}=[x_0, \ldots, x_{t-1}]$ is associated with a true label $y$ and a predicted label $\hat{y}$. 
The underlying goal is to identify which parts of the input contribute more toward the prediction $\hat{y}$ \citep{atanasova-etal-2020-diagnostic, wallace-etal-2020-interpreting, madsen-etal-2022-evaluating, chrysostomou-aletras-2022-empirical, lei-etal-2016-rationalizing, chan-etal-2022-unirex, ghasemi-madani-minervini-2023-refer}. 
Most FAs generally fall into gradient, attention, and perturbation-based categories.

\textbf{Gradient-based} methods derive the importance for each token by computing gradients w.r.t. the input \citep{denil2015inputxgradient}. 
%
The resulting gradient 
captures intuitively the \emph{sensitivity} of the model to each element in the input when predicting token $w$. 
While attribution scores are computed for every dimension of input token embeddings, they are generally aggregated at a token level to obtain a more intuitive overview of the influence of individual tokens.

Building upon this, \citet{denil2015inputxgradient} takes the input token vector and multiplies by the gradient (Input$\times$Gradient), while \citet{sundararajan2017integratedgradients} compares the input with a null baseline input when computing the gradients w.r.t. the input (Integrated Gradients). 
\citet{Nielsen_2022} offers a comprehensive overview of other propagation-based FAs.

\textbf{Attention-based} methods are applied to models that include an attention mechanism to weigh the input tokens. 
The assumption is that the attention weights represent the importance of each token. 
These FAs include scaling the attention weights by their gradients, taking the attention scores from the last layer, and recursively computing the attention in each layer \citep{serrano-smith-2019-attention, jain2020attention, abnar2020rollout}.

\textbf{Perturbation-based methods} measure the difference in model prediction between using the original input and a corrupted version of the input by gradually removing tokens \citep{lei-etal-2016-rationalizing, nguyen-2018-comparing, bastings-etal-2019-interpretable, bashier-etal-2020-rancc}. 
The underlying idea is that removing important tokens will lead the model to flip its prediction or a significant drop in the prediction confidence.
For instance, the input token at position $i$ can be removed, and the resulting probability difference $\mathcal{F}_\theta(\text{X}) - \mathcal{F}_\theta(\mathbf{\text{X}}\setminus x_i)$ can be used as an estimate for its importance. If the logit or probability given to the original output does not change, we conclude that the $i$-th token has no influence.
Differently, some perturbation-based techniques utilize a modified model or a separate explainer model to learn feature attributions \citep{ribeiro2016lime, lundberg2017gradientshap, bashier-etal-2020-rancc, hase2021outofdist}. LIME \citep{ribeiro2016lime} and  SHAP \citep{lundberg2017gradientshap} fall into this category.

\section{Conclusion}
In this paper, we introduced \textsc{Noiser}, a perturbation-based input attribution method that employs bounded noise to address the distribution shift problem arising from the discrete nature of text, aiming to explain language model predictions in generation tasks. Furthermore, we proposed \emph{answerability} metrics, a novel automatic plausibility evaluation metric that leverages an LLM to evaluate the relevance of attributed rationales to the target output in the absence of gold rationales or human evaluation.
Through comprehensive experiments across three tasks and six LLMs, we demonstrated that \textsc{Noiser} consistently surpasses existing baselines in terms of both faithfulness and answerability rate. Notably, our approach requires no supervision, positioning it as a promising direction for improving model interpretability and efficiency. 

%
%
%
%
%
%

\bibliography{colm2025_conference, anthology}
\bibliographystyle{colm2025_conference}

\appendix



\section{Soft-NC and Soft-NS Metrics}
\label{sec:metrics}
The Soft-NC and Soft-NS metrics are defined as follows:

\begin{align}
    \label{eq: soft-c}
    \text{Soft-C}(\text{X}, \hat{y}, \text{X}') &= \max(0, p(\hat{y} \mid \text{X}) - p(\hat{y} \mid \text{X}')) \\
    \label{eq: soft-s}
    \text{Soft-S}(\text{X}, \hat{y}, \text{X}') &= 1 - \text{Soft-C}(\text{X}, \hat{y}, \text{X}')
\end{align}
where $\text{X}'$ is soft-perturbed versions of $\text{X}$ given the following instruction. 
For the embedding vector $\text{x}_i \in \text{X}$ and its FA score $s_i$, we modify the elements of $\text{x}_i$ using \cref{eq: purtubing}.
\begin{align}
    \label{eq: purtubing}
    \text{x}_i' = \text{x}_i \odot \text{e}_i, \quad \text{e}_i \sim \text{Bernoulli}(q)
\end{align}
where $\text{e}$ is a binary mask vector of size $n$ (embedding size) and Bernoulli is parameterized with probability $q$:
\begin{align}
\label{eq: q}
    q =
    \begin{cases}
        s, & \text{if retaining elements} \\
        1 - s, & \text{if removing elements}
    \end{cases}
\end{align}
The normalized sufficiency and comprehensiveness are then computed using the following equations:

\begin{align}
    \text{Soft-NC}(\text{X}, \hat{y}, \text{X}') &= \frac{\text{Soft-C}(\text{X}, \hat{y}, \text{X}')}{1 - S(\text{X}, \hat{y}, 0)} \\
    \text{Soft-NS}(\text{X}, \hat{y}, \text{X}') &= \frac{\text{Soft-S}(\text{X}, \hat{y}, \text{X}') - S(\text{X}, \hat{y}, 0)}{1 - S(\text{X}, \hat{y}, 0)}
\end{align}

However, in generation tasks, the absence of a predictive likelihood for the predicted label makes applying Soft-NS and Soft-NC challenging. 
\citet{zhao-aletras-2023-incorporating} proposed using the Hellinger distance between prediction distributions over the vocabulary as a measure of changes in model predictions. 
They substitute \( p(\hat{y} \mid \text{X}) - p(\hat{y} \mid \text{X}') \) in \cref{eq: soft-c} with the Hellinger distance. 
Given two discrete probability distributions, \( P_{\text{X},t} = [p_{1,t}, \ldots, p_{v,t}] \) and \( P_{\text{X}',t} = [p'_{1,t}, \ldots, p'_{v,t}] \), the Hellinger distance is formally defined as:

\begin{align*}
    \Delta P_{\text{X}',t} &= H(P_{\text{X},t}, P_{\text{X}',t}) = \frac{1}{\sqrt{2}}\cdot\sqrt{\sum_{i=1}^{v} \left( \sqrt{p_{i,t}} - \sqrt{p'_{i,t}} \right)^2}
\end{align*}

where $P_{\text{X},t}$ is the probability distribution over the entire vocabulary (of size $v$) when prompting the model with the full-text $\text{X}$. $P_{\text{X}',t}$ is for prompting the model with soft-perturbed text. 
For a given sequence input $\text{X}$ and a model of vocabulary size $v$, at time step $t$, the model generates a distribution $P_{\text{X},t}$ for the next token $x_T$. 
The final Soft-NS and Soft-NC at step $t$ for text generation are formulated as:

\begin{align}
    \text{Soft-NS}(\text{X}, x_t, \mathcal{R}) &= \frac{\max(0, \Delta P_{\text{0},t} - \Delta P_{\text{X}',t})}{\Delta P_{\text{0},t}} \\
    \text{Soft-NC}(\text{X}, x_t, \mathcal{R}) &= \frac{\Delta P_{\text{X}' \setminus \mathcal{R}, t}}{\Delta P_{\text{0},t}}
\end{align}

where $\Delta P_{\text{0},t}$ is Hellinger's distance between a zero input's probability distribution and full-text input's probability distribution. $\text{X}' \setminus \mathcal{R}$ is the case of ``if removing elements'' described in \cref{eq: q}.

\section{More Results}
\Cref{tab:soft-ns-nc} provides the detailed Soft-NS and Soft-NC scores of our experiments.

\input{tables/soft-ns-nc}

\Cref{tab:answerability-top5} illustrates the answerability metrics on top-50\% of feature attributions given the judge model top-5 predictions.
\input{tables/answerability-top5}

\begin{figure}[h]
    \centering
    \begin{subfigure}[b]{0.4\linewidth}
        \centering
        \includegraphics[width=\linewidth]{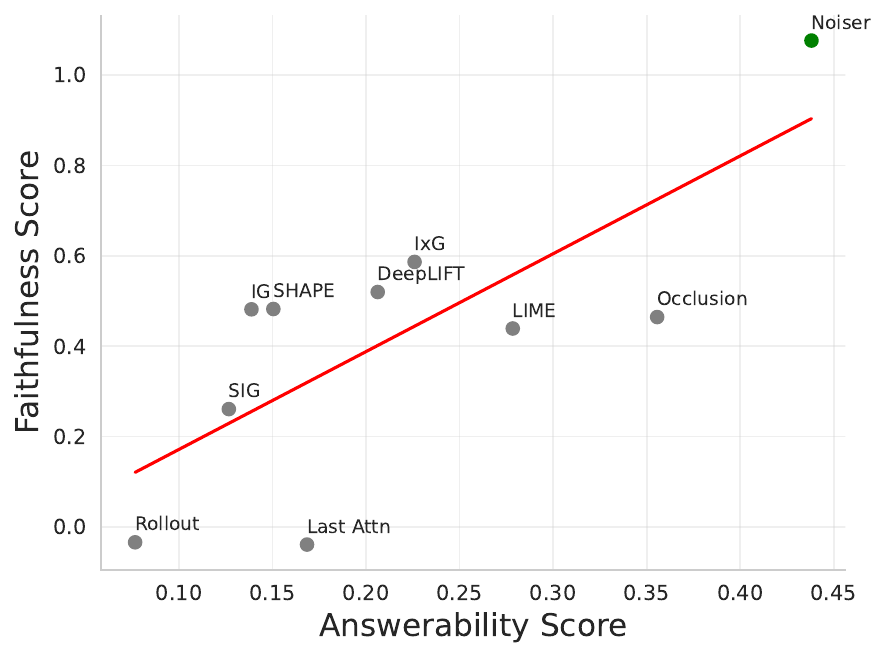}
        \caption{}
        \label{fig:faithVSscore}
    \end{subfigure}
    \begin{subfigure}[b]{0.4\linewidth}
        \centering
        \includegraphics[width=\linewidth]{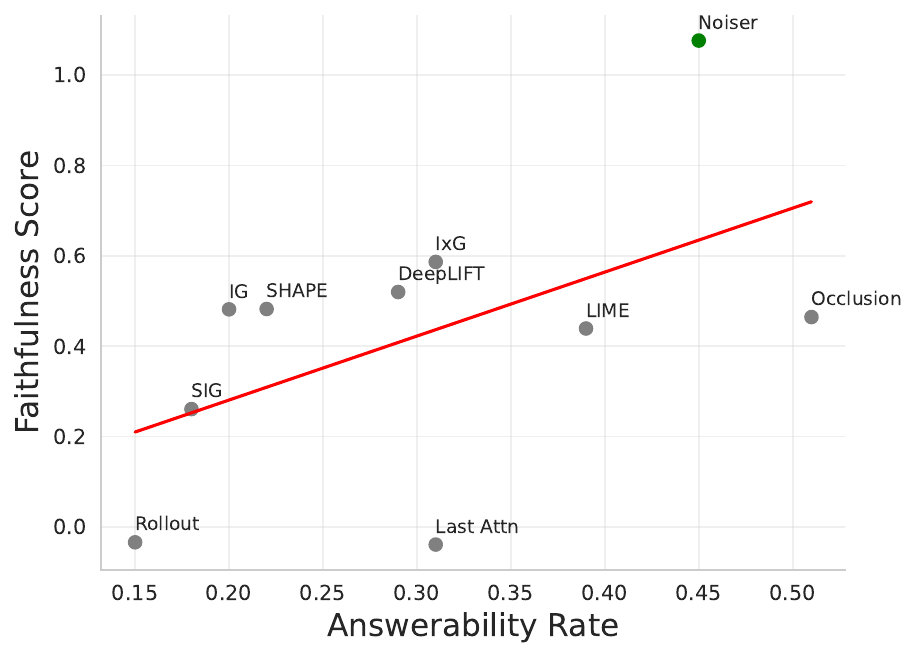}
        \caption{}
        \label{fig:faithVSrate}
    \end{subfigure}
    \caption{Comparison of average faithfulness score with (a) average answerability score and (b) answerability rate.}
    \label{fig:faith-ans}
\end{figure}

We also analyzed the correlation between the faithfulness score and the answerability metrics on \textsc{Known} in \cref{fig:faith-ans}. The results show that IG, SHAP, DeepLIFT, IxG, LIME, and Occlusion exhibit similar faithfulness scores, with Occlusion achieving the highest answerability rate and score among them. Meanwhile, \textsc{Noiser} surpasses all baselines in both faithfulness and answerability score but falls short in answerability rate, trailing Occlusion by 5\%.

\section{Datasets Statistics}
\label{sec:data-stat}
\cref{tab:data-stat} shows the number of true predictions by each model given \textsc{Known} and \textsc{LongRA} datasets.

\begin{table*}[h]
\centering

\label{tab:data-stat}
\resizebox{\textwidth}{!}{%
\begin{tabular}{rcccccc}
\toprule
\textbf{Dataset} & \textbf{Qwen2-0.5B} & \textbf{Llama3.2-1b} & \textbf{Qwen2-1.5B} & \textbf{gemma-2-2b} & \textbf{gemma-2-9b} & \textbf{Llama3-8b} \\ \midrule
Knowns (1208)  & 661        & 828         & 774        & 830        & 822      & 875 \\
LongRA (573)  & 140        & 160         & 170        & 209        & 148      & 165 \\ \bottomrule
\end{tabular}%
}
\caption{Number of true predictions captured by each model.}
\end{table*}


\section{Bounding Computations}
\label{sec:scaling-effect}
Since $k_{\min}$ is dependent on the model, we introduce norm-bounding as the norm value of the noise vector $\mathbf{n}$ is different based on the model's embedding size ($d_{model}$). To avoid different norm values for each sample in a given data, we use the expected norm value of the noise vector, $\mathbb{E}\left[\|\mathbf{n}\|_p\right]$, and use $k=\frac{1}{\mathbb{E}\left[\|\mathbf{n}\|_p\right]}$ as the final bounding for the noise vector. In the following, we show the expected value of each norm given a model with $d_{\text{model}}$ embedding dimensions.

Let $\mathbf{n} \in \mathbb{R}^{d_{\text{model}}}$ be a random vector where each component $n_i \sim \mathcal{N}(0, 1)$. Below, we derive the expected values of different norms and compare their properties.

The $L_2$ norm (Euclidean Norm) is defined as follows:
\begin{equation*}
    \|\mathbf{n}\|_2 = \sqrt{\sum_{i=1}^{d_{\text{model}}} n_i^2}
\end{equation*}
where each $n_i^2$ follows a \textit{chi-squared distribution} with 1 degree of freedom, which results in the following:
\begin{equation*}
    \mathbb{E}\left[\|\mathbf{n}\|_2^2\right] = d_{\text{model}}
\end{equation*}
By Jensen’s inequality and the Law of Large Numbers, for large $d_{\text{model}}$:
\begin{equation*}
    \mathbb{E}\left[\|\mathbf{n}\|_2\right] \approx \sqrt{\mathbb{E}\left[\|\mathbf{n}\|_2^2\right]} = \sqrt{d_{\text{model}}}
\end{equation*}

The $L_\infty$ norm (Maximum Norm) is defined as follows:
\begin{equation*}
    \|\mathbf{n}\|_\infty = \max_{1 \leq i \leq d_{\text{model}}} |n_i|
\end{equation*}
The cumulative distribution function (CDF) for $|n_i|$ is $F(x) = \text{erf}\left(\frac{x}{\sqrt{2}}\right)$. The CDF for the maximum of $d_{\text{model}}$ samples is $F_{\text{max}}(x) = \left[F(x)\right]^{d_{\text{model}}}$. Using extreme value theory, the expected maximum for large $d_{\text{model}}$ approximates:
\begin{equation*}
    \mathbb{E}\left[\|\mathbf{n}\|_\infty\right] \approx \sqrt{2 \ln d_{\text{model}}}
\end{equation*}


\section{Answerability Evaluation Prompt}
\label{sec:ans-prompt}
Below, we provide the prompt used for evaluating FAs' answerability. 

\begin{tcolorbox}[
        colback=gray!25!white,
        colframe=gray!100!gray,
        colbacktitle=gray!100!white, 
        title=Answerability Evaluation Prompt,
        fonttitle=\bfseries, 
        width=\textwidth
        ]
\small
\label{plaus_prompt}
\# Task:\\
Given a set of words extracted from a prompt for a completion task, return a single word as the most probable completion for the unseen prompt WITHOUT providing any explanation.
\end{tcolorbox}

\end{document}

%% file: tables/faithfulness-all.tex
\begin{table}[t]
    \centering
    \begin{subtable}{\textwidth}
        \centering
        \caption*{\textsc{Knowns}}
        \label{tab:faithfulness-knowns}
        \resizebox{\textwidth}{!}{%
        \begin{tabular}{rcccccc|c} 
            \toprule
            \textbf{Method}
            & \textbf{Qwen2-0.5B} & \textbf{Llama3.2-1B} & \textbf{Qwen2-1.5B} & \textbf{gemma-2-2B} & \textbf{gemma-2-9B} & \textbf{Llama3-8B} & \textbf{Average} \\
            \midrule
            Last Attn & -0.0857 & -0.0601 & 0.0607 & 0.1407 & -0.2788 & -0.0095 & -0.0388\\
            Rollout  & -0.1161 & -0.0471 & 0.1211 & 0.4624 & -0.3607 & -0.2625 & -0.0338\\
            SHAP  & 0.4946 & 0.3746 & 0.5390 & 0.3726 & 0.9203 & 0.1925 & 0.4823\\
            IxG  & 0.2117 & 0.7059 & 0.4612 & 0.5233 & 1.0276 & 0.5891 & 0.5865\\
            IG  & 0.2176 & 0.5428 & 0.5163 & 0.2015 & 1.0355 & 0.3759 & 0.4816\\
            DeepLIFT  & 0.3030 & 0.5473 & 0.5323 & 0.3557 & 0.8638 & 0.5174 & 0.5199\\
            SIG  & 0.0361 & 0.3534 & 0.3003 & -0.1879 & 0.7877 & 0.2755 & 0.2609\\
            LIME  & 0.2439 & 0.5103 & 0.3826 & 0.3567 & 0.4832 & 0.6555 & 0.4392\\
            Occlusion  & 0.1627 & 0.5373 & 0.2477 & 0.5341 & 0.5221 & 0.7831 & 0.4645\\
            \rowcolor[HTML]{C0C0C0} 
            \textsc{\textbf{Noiser}} & \textbf{2.1854} & \textbf{1.3989} & \textbf{1.4400} & \textbf{1.4433} & \textbf{2.1767} & \textbf{2.2175} & \textbf{1.8103}\\
            \bottomrule
        \end{tabular}
        }
    \end{subtable}
    
    \vspace{0.01em} 
    
    \begin{subtable}{\textwidth}
        \centering
        \caption*{\textsc{LongRA}}
        \label{tab:faithfulness-longra}
        \resizebox{\textwidth}{!}{%
        \begin{tabular}{rcccccc|c} 
            \toprule
            \textbf{Method}
            & \textbf{Qwen2-0.5B} & \textbf{Llama3.2-1B} & \textbf{Qwen2-1.5B} & \textbf{gemma-2-2B} & \textbf{gemma-2-9B} & \textbf{Llama3-8B} & \textbf{Average} \\
            \midrule
            Last Attn & 1.9148 & 0.3255 & -0.0110 & -0.2382 & -0.2382 & 1.0762 & 0.4715\\
            Rollout  & 1.8517 & 0.2451 & 0.0802 & -0.2643 & -0.2643 & 1.2283 & 0.4794\\
            SHAP  & 3.7970 & 1.2837 & 1.6276 & 1.9746 & 2.2769 & 0.7696 & 1.9549\\
            IxG  & 3.8972 & 1.7299 & 1.5370 & 2.5803 & 2.5803 & 2.0796 & 2.4007\\
            IG  & 4.3388 & 1.3066 & 1.5498 & 1.3023 & 1.3023 & 3.7190 & 2.2531\\
            DeepLIFT  & 4.4991 & 1.7889 & 1.5512 & 2.7428 & 2.7428 & 2.1258 & 2.5751\\
            SIG  & 3.8645 & 0.9272 & 1.1047 & 0.5412 & 0.5412 & 1.1618 & 1.3568\\
            LIME  & 1.0765 & 0.2212 & -0.4147 & -0.1636 & -0.1636 & 2.2995 & 0.4759\\
            Occlusion  & 3.9424 & 1.9887 & 1.0145 & 3.4418 & 3.4418 & 4.2240 & 3.0089\\
            \rowcolor[HTML]{C0C0C0} 
            \textsc{\textbf{Noiser}} & \textbf{6.8055} & \textbf{4.8072} & \textbf{3.1779} & \textbf{4.2727} & \textbf{6.1681} & \textbf{5.1627} & \textbf{5.0657}\\
            \bottomrule
        \end{tabular}
        }
    \end{subtable}
    
    \vspace{0.01em} 

    \begin{subtable}{\textwidth}
        \centering
        \caption*{\textsc{WikiBio}}
        \label{tab:faithfulness-wikibio}
        \resizebox{\textwidth}{!}{%
        \begin{tabular}{rcccccc|c} 
            \toprule
            \textbf{Method}
            & \textbf{Qwen2-0.5B} & \textbf{Llama3.2-1B} & \textbf{Qwen2-1.5B} & \textbf{gemma-2-2B} & \textbf{gemma-2-9B} & \textbf{Llama3-8B} & \textbf{Average} \\
            \midrule
            Last Attn   & 1.0605 & 0.6304 & -0.7054 & -0.2579 & 0.2815 & 0.5500 & 0.2598 \\
            Rollout     & -0.6404 & 0.5591 & -0.7066 & 0.5085 & 0.3498 & 0.8785 & 0.1582 \\
            SHAP        & 1.4702 & 1.1672 & 1.1213 & 0.7966 & 3.1494 & 1.4063 & 1.5185 \\
            IxG         & 3.4273 & 1.8365 & 1.3942 & 1.5816 & 2.6047 & 1.3747 & 2.0365 \\
            IG          & 2.4216 & 1.5797 & 0.6975 & 1.1909 & 4.1117 & 0.6876 & 1.7815 \\
            DeepLIFT    & 3.2207 & 1.6265 & 1.4590 & 1.4607 & 2.3006 & 1.2739 & 1.8903 \\
            SIG         & 3.7656 & 1.4300 & 2.0816 & 1.4256 & 5.2280 & 1.3620 & 2.5488 \\
            LIME        & 3.0009 & 0.5656 & 1.1714 & 0.7180 & 2.9527 & 0.8349 & 1.5406 \\
            Occlusion   & 5.1051 & 2.0019 & 3.8916 & 2.7232 & 4.9300 & 3.3885 & 3.6734 \\
            \rowcolor[HTML]{C0C0C0} 
            \textsc{\textbf{Noiser}} & \textbf{8.7624} & \textbf{3.7385} & \textbf{4.9864} & \textbf{4.2527} & \textbf{7.1509} & \textbf{4.6089} & \textbf{5.5833} \\
            \bottomrule
        \end{tabular}
        }
    \end{subtable}
    \caption{Faithfulness scores across tasks.}
    \label{tab:faithfulness-all}
\end{table}

%% file: tables/k-effect.tex
\begin{table}[t]
    \centering
    \resizebox{\textwidth}{!}{%
    \begin{tabular}{lcccccc|c} 
        \toprule
        \textbf{Scaling Factor ($k$)}
        & \textbf{Qwen2-0.5B} & \textbf{Llama3.2-1B} & \textbf{Qwen2-1.5B} & \textbf{gemma-2-2B} & \textbf{gemma-2-9B} & \textbf{Llama3-8B} & \textbf{Average} \\
        \midrule
        random $k$ & 1.1519 & 1.0993 & 0.7165 & 1.2287 & 1.6007 & 1.4522 & 1.2082 \\
        None ($k=1$) & 1.0922 & 1.0219 & 0.6445 & 1.1844 & 1.4726 & 1.1070 & 1.0871 \\
        $\mathbb{E}\left[\|\mathbf{n}\|_2\right]^{-1}$  & 1.4849 & 1.3031 & 1.2617 & 1.7236 & 2.7905 & 1.7470 & 1.7185 \\
        $\mathbb{E}\left[\|\mathbf{n}\|_\infty\right]^{-1}$  & 0.8989 & 0.9515 & 0.7164 & 1.1275 & 1.4300 & 1.4115 & 1.0893 \\
        $k_{\max}$ per token & 1.2230 & 0.9938 & 0.5850 & 1.2897 & 1.8359 & 1.0984 & 1.1710 \\
        $k_{\max}$ & 1.0962 & 1.0203 & 0.6515 & 1.1824 & 1.4753 & 1.1059 & 1.0886 \\
        \rowcolor[HTML]{C0C0C0} 
        $k_{\min}$ & \textbf{2.1854} & \textbf{1.3989} & \textbf{1.4400} & 1.4433 & 2.1767 & \textbf{2.2175} & \textbf{1.8103} \\
        \bottomrule
    \end{tabular}
    }    
    \caption{Comparison of different boundings on the faithfulness score on \textsc{Known} dataset.}    \label{tab:k-effect}
\end{table}

%% file: tables/answerability-top1.tex
\begin{table}[t]
    \centering
    \resizebox{\textwidth}{!}{%
    \begin{tabular}{rcc|cc|cc|cc|cc|cc|cc}
        \toprule
        \multirow{2}{*}{\textbf{Method}}
        & \multicolumn{2}{c}{\textbf{Qwen2-0.5B}} & \multicolumn{2}{c}{\textbf{Llama3.2-1b}} & \multicolumn{2}{c}{\textbf{Qwen2-1.5B}} & \multicolumn{2}{c}{\textbf{gemma-2-2b}} & \multicolumn{2}{c}{\textbf{gemma-2-9b}} & \multicolumn{2}{c}{\textbf{Llama3-8b}} & \multicolumn{2}{c}{\textbf{Average}} \\
        \cmidrule(lr){2-3} \cmidrule(lr){4-5} \cmidrule(lr){6-7} \cmidrule(lr){8-9} \cmidrule(lr){10-11} \cmidrule(lr){12-13} \cmidrule(lr){14-15}
        & \textbf{Rate} & \textbf{Score} & \textbf{Rate} & \textbf{Score} & \textbf{Rate} & \textbf{Score} & \textbf{Rate} & \textbf{Score} & \textbf{Rate} & \textbf{Score} & \textbf{Rate} & \textbf{Score} & \textbf{Rate} & \textbf{Score} \\
        \midrule
        Last Attn & 14\% & 0.0936 & 48\% & 0.2496 & 10\% & 0.0670 & 39\% & 0.2064 & 37\% & 0.1854 & 39\% & 0.2081 & 31\% & 0.1684\\
        Rollout & 8\% & 0.0527 & 13\% & 0.0649 & 8\% & 0.0557 & 9\% & 0.0457 & 22\% & 0.1033 & 27\% & 0.1364 & 16\% & 0.0812\\
        SHAP & 22\% & 0.1805 & 29\% & 0.1890 & 24\% & 0.1862 & 17\% & 0.1249 & 11\% & 0.0764 & 26\% & 0.1454 & 22\% & 0.1504\\
        IxG & 27\% & 0.2177 & 33\% & 0.2412 & 26\% & 0.1942 & 35\% & 0.2408 & 35\% & 0.2537 & 30\% & 0.2079 & 31\% & 0.2259\\
        IG & 20\% & 0.1638 & 28\% & 0.1827 & 18\% & 0.1426 & 16\% & 0.1197 & 12\% & 0.0875 & 27\% & 0.1360 & 20\% & 0.1387\\
        DeepLIFT & 21\% & 0.1753 & 34\% & 0.2279 & 26\% & 0.1991 & 32\% & 0.2225 & 30\% & 0.2107 & 31\% & 0.2019 & 29\% & 0.2062\\
        SIG & 21\% & 0.1583 & 21\% & 0.1271 & 20\% & 0.1520 & 9\% & 0.0617 & 26\% & 0.1978 & 12\% & 0.0627 & 18\% & 0.1266\\
        LIME & 37\% & 0.2986 & 25\% & 0.1692 & 41\% & 0.3308 & 45\% & 0.3003 & 50\% & 0.3291 & 36\% & 0.2423 & 39\% & 0.2784 \\
        Occlusion & 53\% & 0.3689 & \textbf{49\%} & 0.3223 & \textbf{54\%} & \textbf{0.4224} & \textbf{48\%} & 0.3152 & \textbf{52\%} & 0.3323 & \textbf{50\%} & 0.3726 & \textbf{51\%} & 0.3556 \\
        \rowcolor[HTML]{C0C0C0} 
        \textsc{\textbf{Noiser}} & \textbf{55\%} & \textbf{0.5063} & 37\% & \textbf{0.3665} & 43\% & 0.4099 & 43\% & \textbf{0.4102} & 49\% & \textbf{0.4497} & 41\% & \textbf{0.4858} & 45\% & \textbf{0.4381} \\

        \bottomrule
    \end{tabular}
    }
    \caption{Answerability metrics on \textsc{Known} dataset w.r.t. judge model top-1 predition.}
    \label{tab:answerability}
\end{table}

%% file: tables/highlights.tex
\begin{table}[h]
    \small
    \centering
        \begin{tabularx}{\textwidth}{l|X|l}
        \toprule
        \textbf{Dataset} & \textbf{Input} & \textbf{Output} \\ \midrule
        Knowns  & \hlc{teal!100.000}{LeBron} \hlc{teal!6.743}{James} \hlc{teal!44.172}{professionally} \hlc{teal!7.759}{plays} \hlc{teal!0.000}{the} \hlc{teal!96.544}{sport} \hlc{teal!41.553}{of} & basketball \\ \midrule
        LongRA  &  \hlc{teal!8.112}{When} \hlc{teal!10.401}{my} \hlc{teal!16.592}{flight} \hlc{teal!18.370}{landed} \hlc{teal!9.173}{in} \hlc{teal!84.461}{Japan} \hlc{teal!8.914}{,} \hlc{teal!9.886}{I} \hlc{teal!6.707}{converted} \hlc{teal!8.400}{my} \hlc{teal!14.819}{currency} \hlc{teal!10.616}{and} \hlc{teal!13.835}{slowly} \hlc{teal!6.717}{fell} \hlc{teal!10.327}{asleep} \hlc{teal!7.831}{.} \hlc{teal!5.686}{(} \hlc{teal!9.366}{I} \hlc{teal!8.183}{had} \hlc{teal!7.937}{a} \hlc{teal!8.327}{terrifying} \hlc{teal!10.705}{dream} \hlc{teal!7.938}{about} \hlc{teal!8.354}{my} \hlc{teal!11.528}{grandmother} \hlc{teal!7.415}{,} \hlc{teal!7.614}{but} \hlc{teal!8.378}{that} \hlc{teal!10.088}{’} \hlc{teal!10.088}{s} \hlc{teal!7.619}{a} \hlc{teal!6.335}{story} \hlc{teal!8.235}{for} \hlc{teal!11.296}{another} \hlc{teal!10.211}{time} \hlc{teal!10.238}{)} \hlc{teal!10.238}{.} \hlc{teal!6.289}{I} \hlc{teal!6.413}{was} \hlc{teal!12.517}{staying} \hlc{teal!16.608}{in} \hlc{teal!8.438}{the} \hlc{teal!100.000}{capital} \hlc{teal!0.000}{,} & Tokyo \\ \midrule
        WikiBio &\hlc{teal!3.870}{Super} \hlc{teal!100.000}{Mario} \hlc{teal!22.750}{Land} \hlc{teal!1.390}{is} \hlc{teal!0.451}{a} \hlc{teal!1.868}{1989} \hlc{teal!2.030}{side-scrolling} \hlc{teal!0.000}{platform} \hlc{teal!0.164}{video} \hlc{teal!0.349}{game} \hlc{teal!0.382}{developed} \hlc{teal!0.115}{and} \hlc{teal!9.326}{published} \hlc{teal!74.404}{by} & Nintendo \\ \bottomrule
        \end{tabularx}
    \caption{Example of \textsc{Noiser} attributions on different inputs.}
    \label{tab:highlights}
\end{table}

%% file: tables/soft-ns-nc.tex
\begin{table}[t]
    \centering
    \begin{subtable}{\textwidth}
        \centering
        \caption*{\textsc{Knowns}}
        \resizebox{\textwidth}{!}{%
        \begin{tabular}{rcc|cc|cc|cc|cc|cc}
            \toprule
            \multirow{2}{*}{\textbf{Method}}
            & \multicolumn{2}{c}{\textbf{Qwen2-0.5B}} & \multicolumn{2}{c}{\textbf{Llama3.2-1b}} & \multicolumn{2}{c}{\textbf{Qwen2-1.5B}} & \multicolumn{2}{c}{\textbf{gemma-2-2b}} & \multicolumn{2}{c}{\textbf{gemma-2-9b}} & \multicolumn{2}{c}{\textbf{Llama3-8b}} \\
            \cmidrule(lr){2-3} \cmidrule(lr){4-5} \cmidrule(lr){6-7} \cmidrule(lr){8-9} \cmidrule(lr){10-11} \cmidrule(lr){12-13}
            & \textbf{Soft-NS} & \textbf{Soft-NC} & \textbf{Soft-NS} & \textbf{Soft-NC} & \textbf{Soft-NS} & \textbf{Soft-NC} & \textbf{Soft-NS} & \textbf{Soft-NC} & \textbf{Soft-NS} & \textbf{Soft-NC} & \textbf{Soft-NS} & \textbf{Soft-NC} \\
            \midrule
            Last Attn   & 0.0372 & -0.1229 & -0.0301 & -0.0301 & 0.0023 & 0.0584 & 0.1617 & -0.0211 & 0.0321 & -0.3109 & -0.1148 & 0.1052  \\
            Rollout     & -0.2567 & 0.1406 & -0.0426 & -0.0045 & -0.0264 & 0.1475 & 0.3582 & 0.1042 & 0.0114 & -0.3721 & -0.5818 & 0.3194  \\
            SHAP        & -0.2714 & 0.7660 & -0.1643 & 0.5388 & -0.0809 & 0.6199 & -0.0050 & 0.3776 & 0.1933 & 0.7270 & -0.3386 & 0.5310  \\
            IxG         & -0.5373 & 0.7490 & 0.0079 & 0.6980 & -0.0843 & 0.5455 & -0.1152 & 0.6384 & -0.0265 & 1.0541 & 0.0750 & 0.5141  \\
            IG          & -0.6136 & 0.8312 & -0.1202 & 0.6629 & -0.1209 & 0.6372 & -0.1328 & 0.3343 & 0.1540 & 0.8815 & -0.3386 & 0.7144  \\
            DeepLIFT    & -0.5430 & 0.8460 & -0.0801 & 0.6274 & -0.0826 & 0.6149 & -0.1296 & 0.4853 & -0.0109 & 0.8747 & 0.0147 & 0.5027\\
            SIG         & -0.4989 & 0.5350 & -0.0841 & 0.4375 & -0.0964 & 0.3969 & -0.1494 & -0.0385 & 0.1730 & 0.6147 & -0.1886 & 0.4640  \\
            LIME        & -0.0221 & 0.2660 & 0.2210 & 0.2920 & 0.0678 & 0.3149 & 0.1336 & 0.2230 & 0.2100 & 0.2732 & 0.5459 & 0.1096  \\
            Occlusion   &  0.1138 & 0.0489 & 0.2267 & 0.3105 & 0.0606 & 0.1872 & 0.1830 & 0.3510 & 0.2119 & 0.3102 & 0.8302 & -0.0470  \\
            \rowcolor[HTML]{C0C0C0}
            \textsc{\textbf{Noiser}} &  0.6785 & 1.5068 & 0.2248 & 1.1741 & -0.0264 & 1.4664 & 0.1805 & 1.2627 & -0.0233 & 2.2001 & 0.8043 & 1.4133\\
            \bottomrule
        \end{tabular}
        }
    \end{subtable}
    
    \vspace{0.01em} 
    
    \begin{subtable}{\textwidth}
        \centering
        \caption*{\textsc{LongRA}}
        \resizebox{\textwidth}{!}{%
        \begin{tabular}{rcc|cc|cc|cc|cc|cc}
            \toprule
            \multirow{2}{*}{\textbf{Method}}
            & \multicolumn{2}{c}{\textbf{Qwen2-0.5B}} & \multicolumn{2}{c}{\textbf{Llama3.2-1b}} & \multicolumn{2}{c}{\textbf{Qwen2-1.5B}} & \multicolumn{2}{c}{\textbf{gemma-2-2b}} & \multicolumn{2}{c}{\textbf{gemma-2-9b}} & \multicolumn{2}{c}{\textbf{Llama3-8b}} \\
            \cmidrule(lr){2-3} \cmidrule(lr){4-5} \cmidrule(lr){6-7} \cmidrule(lr){8-9} \cmidrule(lr){10-11} \cmidrule(lr){12-13}
            & \textbf{Soft-NS} & \textbf{Soft-NC} & \textbf{Soft-NS} & \textbf{Soft-NC} & \textbf{Soft-NS} & \textbf{Soft-NC} & \textbf{Soft-NS} & \textbf{Soft-NC} & \textbf{Soft-NS} & \textbf{Soft-NC} & \textbf{Soft-NS} & \textbf{Soft-NC} \\
            \midrule
            Last Attn   & 1.8502 & 0.0645 & 0.2522 & 0.0733 & 0.0828 & -0.0938 & -0.1583 & -0.0799 & -0.1583 & -0.0799 & 0.0888 & 0.9874  \\
            Rollout     & 1.8541 & -0.0025 & 0.0894 & 0.1557 & 0.0927 & -0.0126 & -0.1790 & -0.0853 & -0.1790 & -0.0853 & 0.1052 & 1.1231  \\
            SHAP        & 1.5203 & 2.2767 & -0.3437 & 1.6274 & -0.0323 & 1.6599 & 0.1364 & 1.8382 & 0.1444 & 2.1325 & -0.4277 & 1.1972  \\
            IxG         & 1.3174 & 2.5798 & -0.1419 & 1.8718 & -0.0682 & 1.6052 & 0.2919 & 2.2883 & 0.2919 & 2.2883 & 0.1390 & 1.9406  \\
            IG          & 1.5378 & 2.8010 & -0.3141 & 1.6207 & -0.1039 & 1.6537 & 0.3475 & 0.9548 & 0.3475 & 0.9548 & 0.0639 & 3.6551  \\
            DeepLIFT    & 1.6777 & 2.8214 & -0.0546 & 1.8435 & -0.1606 & 1.7118 & 0.2430 & 2.4998 & 0.2430 & 2.4998 & 0.0657 & 2.0600  \\
            SIG         & 2.4897 & 1.3749 & -0.3132 & 1.2404 & -0.0297 & 1.1344 & 0.1009 & 0.4404 & 0.1009 & 0.4404 & 0.1895 & 0.9723  \\
            LIME        & 0.9895 & 0.0870 & 0.1343 & 0.0869 & -0.4042 & -0.0106 & -0.0112 & -0.1524 & -0.0112 & -0.1524 & 0.0923 & 2.2071  \\
            Occlusion   & 2.2561 & 1.6862 & 0.1370 & 1.8517 & -0.4008 & 1.4153 & 0.5531 & 2.8887 & 0.5531 & 2.8887 & -0.0124 & 4.2364  \\
            
            \rowcolor[HTML]{C0C0C0}
            \textsc{\textbf{Noiser}} &  2.0935 & 4.7119 & 0.3242 & 4.4831 & -0.9588 & 4.1367 & 0.5164 & 3.7563 & 1.1446 & 5.0235 & 0.4705 & 4.6922  \\
            \bottomrule
        \end{tabular}
        }
    \end{subtable}
    
    \vspace{0.01em} 

    \begin{subtable}{\textwidth}
        \centering
        \caption*{\textsc{WikiBio}}
        \resizebox{\textwidth}{!}{%
        \begin{tabular}{rcc|cc|cc|cc|cc|cc}
            \toprule
            \multirow{2}{*}{\textbf{Method}}
            & \multicolumn{2}{c}{\textbf{Qwen2-0.5B}} & \multicolumn{2}{c}{\textbf{Llama3.2-1b}} & \multicolumn{2}{c}{\textbf{Qwen2-1.5B}} & \multicolumn{2}{c}{\textbf{gemma-2-2b}} & \multicolumn{2}{c}{\textbf{gemma-2-9b}} & \multicolumn{2}{c}{\textbf{Llama3-8b}} \\
            \cmidrule(lr){2-3} \cmidrule(lr){4-5} \cmidrule(lr){6-7} \cmidrule(lr){8-9} \cmidrule(lr){10-11} \cmidrule(lr){12-13}
            & \textbf{Soft-NS} & \textbf{Soft-NC} & \textbf{Soft-NS} & \textbf{Soft-NC} & \textbf{Soft-NS} & \textbf{Soft-NC} & \textbf{Soft-NS} & \textbf{Soft-NC} & \textbf{Soft-NS} & \textbf{Soft-NC} & \textbf{Soft-NS} & \textbf{Soft-NC} \\
            \midrule
            Last Attn   & 0.8145 & 0.2459 & 0.1041 & 0.5263 & 0.0126 & -0.7181 & -0.1474 & -0.1105 & 0.2246 & 0.0569 & -0.0905 & 0.6405  \\
            Rollout     & -0.6789 & 0.0385 & 0.0254 & 0.5336 & 0.0424 & -0.7490 & -0.0168 & 0.5253 & 0.2655 & 0.0843 & -0.2114 & 1.0899  \\
            SHAP        & 0.6624 & 0.8078 & 0.0470 & 1.1202 & -0.0263 & 1.1476 & 0.0071 & 0.7895 & 0.4254 & 2.7240 & -0.0304 & 1.4368  \\
            IxG         & 1.1577 & 2.2695 & 0.0991 & 1.7374 & 0.2109 & 1.1833 & 0.0320 & 1.5496 & 0.4097 & 2.1950 & 0.2509 & 1.1238  \\
            IG          & 0.5209 & 1.9008 & 0.0047 & 1.5749 & -0.0232 & 0.7207 & -0.0408 & 1.2318 & 0.5993 & 3.5124 & -0.4283 & 1.1159  \\
            DeepLIFT    & 0.9232 & 2.2975 & -0.0490 & 1.6755 & 0.0495 & 1.4096 & 0.0574 & 1.4033 & 0.3253 & 1.9753 & 0.1346 & 1.1393  \\
            SIG         & 1.4616 & 2.3040 & -0.1815 & 1.6114 & 0.3493 & 1.7323 & 0.1572 & 1.2684 & 0.9031 & 4.3250 & 0.6571 & 0.7049  \\
            LIME        & 1.6490 & 1.3519 & 0.1118 & 0.4537 & 0.1616 & 1.0099 & 0.1260 & 0.5920 & 0.7893 & 2.1634 & 0.3183 & 0.5166  \\
            Occlusion   & 2.6100 & 2.4950 & 0.3736 & 1.6283 & 0.9353 & 2.9564 & 0.4595 & 2.2637 & 1.3045 & 3.6255 & 1.5924 & 1.7961  \\
            \rowcolor[HTML]{C0C0C0}
            \textsc{\textbf{Noiser}} &  4.3005 & 4.4619 & 0.6959 & 3.0427 & 0.9847 & 4.0016 & 0.7941 & 3.4586 & 1.5671 & 5.5837 & 1.4975 & 3.1114  \\
            \bottomrule
        \end{tabular}
        }
    \end{subtable}
    \caption{Soft-NS and Soft-NC Scores Across Datasets.}
    \label{tab:soft-ns-nc}
\end{table}

%% file: tables/answerability-top5.tex
\begin{table}[t]
    \centering
    \resizebox{\textwidth}{!}{%
    \begin{tabular}{rcc|cc|cc|cc|cc|cc|cc}
        \toprule
        \multirow{2}{*}{\textbf{Method}}
        & \multicolumn{2}{c}{\textbf{Qwen2-0.5B}} & \multicolumn{2}{c}{\textbf{Llama3.2-1b}} & \multicolumn{2}{c}{\textbf{Qwen2-1.5B}} & \multicolumn{2}{c}{\textbf{gemma-2-2b}} & \multicolumn{2}{c}{\textbf{gemma-2-9b}} & \multicolumn{2}{c}{\textbf{Llama3-8b}} & \multicolumn{2}{c}{\textbf{Average}} \\
        \cmidrule(lr){2-3} \cmidrule(lr){4-5} \cmidrule(lr){6-7} \cmidrule(lr){8-9} \cmidrule(lr){10-11} \cmidrule(lr){12-13} \cmidrule(lr){14-15}
        & \textbf{Rate} & \textbf{Score} & \textbf{Rate} & \textbf{Score} & \textbf{Rate} & \textbf{Score} & \textbf{Rate} & \textbf{Score} & \textbf{Rate} & \textbf{Score} & \textbf{Rate} & \textbf{Score} & \textbf{Rate} & \textbf{Score} \\
        \midrule
        Last Attn   & 26\% & 0.1722 & 58\% & 0.3037 & 21\% & 0.1432 & 55\% & 0.2917 & 51\% & 0.2529 & 48\% & 0.2568 & 43\% & 0.2368\\
        Rollout     & 21\% & 0.1445 & 37\% & 0.1836 & 23\% & 0.1604 & 23\% & 0.1086 & 36\% & 0.1621 & 40\% & 0.1980 & 30\% & 0.1595\\
        SHAP        & 47\% & 0.3904 & 52\% & 0.3440 & 47\% & 0.3662 & 31\% & 0.2188 & 28\% & 0.2051 & 43\% & 0.2427 & 41\% & 0.2945\\
        IxG         & 51\% & 0.4087 & 58\% & 0.4299 & 40\% & 0.3057 & 55\% & 0.3792 & 47\% & 0.3369 & 44\% & 0.3062 & 49\% & 0.3611\\
        IG          & 48\% & 0.4004 & 54\% & 0.3586 & 43\% & 0.3367 & 31\% & 0.2308 & 25\% & 0.1847 & 42\% & 0.2172 & 41\% & 0.2881\\
        DeepLIFT    & 49\% & 0.4138 & 60\% & 0.4067 & 40\% & 0.3159 & 48\% & 0.3271 & 45\% & 0.3164 & 44\% & 0.2898 & 48\% & 0.3450\\
        SIG         & 40\% & 0.3025 & 44\% & 0.2759 & 40\% & 0.3091 & 24\% & 0.1644 & 36\% & 0.2698 & 28\% & 0.1584 & 35\% & 0.2467\\
        LIME        & 49\% & 0.3936 & 49\% & 0.3315 & 53\% & 0.4307 & 61\% & 0.4021 & 60\% & 0.3916 & 48\% & 0.3230 & 53\% & 0.3787\\
        Occlusion   & \textbf{67\%} & 0.4666 & \textbf{61\%} & 0.4001 & \textbf{65\%} & 0.5093 & 64\% & 0.4204 & \textbf{63\%} & 0.4043 & \textbf{57\%} & 0.4241 & \textbf{63\%} & 0.4375\\
        \rowcolor[HTML]{C0C0C0} 
        \textsc{\textbf{Noiser}} & 65\% & \textbf{0.5996} & 58\% & \textbf{0.5435} & 57\% & \textbf{0.5474} & \textbf{66\%} & \textbf{0.6245} & 60\% & \textbf{0.5400} & 54\% & \textbf{0.5962} & 60\% & \textbf{0.5752} \\

        \bottomrule
    \end{tabular}
    }
    \caption{Answerability metrics on \textsc{Known} dataset w.r.t judge model \textbf{top-5} predictions.}
    \label{tab:answerability-top5}
\end{table}